\pdfoutput=1
\documentclass[11pt]{article}

\usepackage{acl}
\usepackage{times}
\usepackage{latexsym}
\usepackage{booktabs} 
\usepackage[T1]{fontenc}
\usepackage[utf8]{inputenc}

\usepackage{microtype}

\usepackage{inconsolata}

\usepackage{graphicx}
\usepackage{multirow}
\usepackage{tikz}
\usepackage{color}
\usepackage{xcolor}
\usepackage{colortbl}
\usepackage{float}
\usepackage{svg}
\usepackage{tabularx}
\usepackage{xcolor}
\usepackage{verbatim}
\usepackage{pgf}
\usepackage{array}
\usepackage{subfigure}

\definecolor{col_table}{HTML}{249BFF}
\usepackage{xspace}

\usepackage{xcolor}
\usepackage{hhline}
\title{Putting Context in Context: \\the Impact of Discussion Structure on Text Classification}

\author{Nicol\`o Penzo$^{1, 2}$, Antonio Longa$^{2}$, Bruno Lepri$^1$, Sara Tonelli$^1$, Marco Guerini$^1$\\
  $^1$ Fondazione Bruno Kessler, Italy\\
  $^2$ University of Trento, Italy \\
  \texttt{\{npenzo,lepri,satonelli,guerini\}@fbk.eu}\\
  \texttt{antonio.longa@unitn.it}}

\begin{document}

\maketitle
\begin{abstract}

Current text classification approaches usually focus on the content to be classified. Contextual aspects (both linguistic and extra-linguistic) are usually neglected, even in tasks based on online discussions. Still in many cases the multi-party and multi-turn nature of the context from which these elements are selected can be fruitfully exploited.  In this work, we propose a series of experiments on a large dataset for stance detection in English, in which we evaluate the contribution of different types of contextual information, i.e. linguistic, structural and temporal, by feeding them as natural language input into a transformer-based model. We also experiment with different amounts of training data and analyse the topology of local discussion networks in a  privacy-compliant way. Results show that  structural information can be highly beneficial to text classification but only under certain circumstances (e.g. depending on the amount of training data and on discussion chain complexity). Indeed, we show that contextual information on smaller datasets from other classification tasks does not yield significant improvements.
Our framework, based on local discussion networks, allows the integration of structural information, while minimising user profiling, thus preserving their privacy.

\end{abstract}

\section{Introduction} 
 
Online conversations are a main channel through which phenomena such as fake news, rumors and hate speech can spread
\cite{sheth2022defining}, political leaning is expressed  
\cite{garimella2018political} and one's health conditions can be revealed 
\cite{guntuku2017detecting}.
All these phenomena can be captured to some degree automatically, 
provided that we have reliable NLP systems able to classify the content of the messages. Most classification approaches focus on the textual content of single comments (or a pair, in the case of stance detection), however little has been done to include the full context of the conversation and test its usefulness in classification tasks.

\begin{figure}[t!]
    \centering
    \includegraphics[]{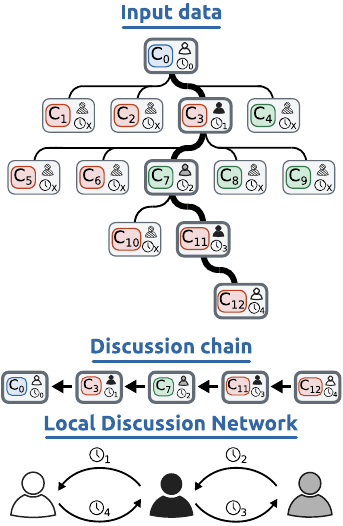}
    \caption{Representation of input data in Kialo dataset: the discussion chain (in bold) is extracted from the discussion tree, and each claim has a textual content $c$, a user id and a timestamp. A \textit{support} (green) or \textit{contrast} (red) label w.r.t. the previous statement is assigned to each claim. The initial claim $c_0$ has no stance (blue).
    This representation can be easily generalized to experiments on other datasets.}
    \label{fig:enter-label}
\end{figure}

Indeed, while the actual content of comments gives us information about what was written, knowing whether and how often two users interact with each other can give us a wider picture of how the dialogue is evolving.
Furthermore, temporal information allows us to identify peaks or  ``waves'' of comments, suggesting the occurrence of a triggering event, as seen in relation to online toxicity \cite{saveski2021structure} and fake news \cite{Vosoughi2018TheSO}.

Previous NLP studies already investigated how contextual information can be included in the classification of online conversations, mainly following three  distinct directions: integrating \textit{textual context}, i.e. the previous thread of a given post \cite{pavlopoulos-etal-2020-toxicity}, modelling \textit{user-related context} \cite{zhang-etal-2018-conversations,nguyen2020fang}, or including \textit{structural context} in terms of conversation structure \cite{song2021temporally, tian-etal-2022-duck}, or external knowledge \cite{beck-etal-2023-robust}.   
Regardless of which type of context was considered, one major issue is represented by the limited size of many benchmarks, from which models can hardly learn contextual information  \cite{menini2021abuse,anuchitanukul2022revisiting}. Another drawback is that, in order to develop classification models embedding contextual information, complex and computationally-intensive architectures are needed \cite{agarwal2022graphnli}. 

We address the above challenges by proposing an approach integrating \textit{textual}, \textit{temporal}  and \textit{structural context} in a simple, unified architecture, where such information is expressed in natural language and is captured by a 
transformer-based model \citep{vaswani2017attention} for classification, without separately modelling the latent structural information of the interactions. In this framework, we avoid to explicitly provide user-related information, which may lead to privacy issues, but we rather represent users as  ``local discussion IDs'', meaning that a user is assigned a new ID for each discussion they participate in. As a consequence, if a user is active in several discussions, this information is not available and user profiling at global network level is not possible, thus enforcing privacy preservation. 

Since previous studies highlighted that training size is crucial to make models aware of contextual information, we mainly perform our experiments on a task of stance detection using a large dataset extracted from the Kialo platform \cite{scialom-etal-2020-toward}. While the dataset is described in detail in Section \ref{sec:kialo}, we report in Figure \ref{fig:enter-label} an example of discussion structure taken from this resource. 

To better understand the contribution of the training set size, we perform also an analysis of the learning curve (Section \ref{sec:second_experiment}) and we evaluate the performance of our models on local discussion networks (LDNs) of different complexity and of varying length (Section \ref{sec:last_experiment}). As a comparison, we also test our approach on two smaller datasets for stance detection and abusive language detection, confirming the effect of dataset size (Section \ref{sec:first_experiment}). 

The data are available upon request only for research purposes, in compliance with Kialo's terms of service. We follow a data minimisation principle, sharing only the information needed to replicate our experiments after user anonymisation.\footnote{The request form and the software to reproduce the experiments are publicly available at  \url{https://github.com/dhfbk/PuCC}.}

\section{Related Work}\label{sec_:related_work}

Despite the fact that social network discussions involve more information than just a sequence of texts, such as user interactions and temporal evolution, researchers have only made few attempts to combine linguistic information with structural and temporal information. Some attempts have been made for tasks like fake news detection (e.g., \citealp{nguyen2020fang}, and \citealp{song2021temporally}), hate speech detection \citep{chakraborty-etal-2022-crush}, stance detection (e.g., \citealp{yang-etal-2019-blcu}, and \citealp{zhou2023stance})
and rumour verification \citep{zhou-etal-2019-early}. User-related information has also been successfully exploited in  abusive comment moderation \cite{pavlopoulos-etal-2017-improved}.

\par All these tasks are closely related to the dynamics of human behavior, but the involvement of linguistic information, network information and temporal information altogether has been difficult because of: \textsc{i.} the fusion of heterogeneous knowledge, by combining computationally-expensive models such as Pretrained Language Models
and Graph Neural Networks (GNNs) \cite{zhou2020graph}, like in \citet{lin-etal-2021-bertgcn}; \textsc{ii.} the access to large-scale private data, that cannot be freely released; \textsc{iii.} the training of human annotators on this data; \textsc{iv.} the deletion of social media posts over time, leading to gaps in discussions, especially in hate speech and fake news datasets \cite{klubicka2018examining}. For few shared tasks, datasets that also include contextual information such as user ids and timestamps have been created \cite{gorrell-etal-2019-semeval,cignarella2020sardistance}. 
Still, researchers have mostly worked only on the textual content.

One of the reasons why contextual information has been marginally explored in classification tasks is that it has not been proved beneficial in a consistent way. As shown by \citet{menini2021abuse}, exploiting the textual context does not lead to any increase in performance for abusive language detection, even if the dataset was re-annotated by looking at the full context. These results have been confirmed by \citet{anuchitanukul2022revisiting}, who further show that the outcome of contextual models strongly depends on the intrinsic characteristics and the dimension of the training set. \citet{yu-etal-2022-hate} show that adding a short context (only parent and target comments) improves hate speech classification. However, they do not consider any structural context but only textual one. 
Similar to our work, \citet{beck-etal-2023-robust} model contextual information through natural language. However, they consider as ``context'' external contextual knowledge such as structured knowledge bases, causal relationships, or information retrieved from a large pretrained model, and not the conversation structure.

As regards stance detection, \citet{agarwal2022graphnli} propose a graph-based inference model to predict the stance of a comment versus its own parent, exploiting the concept of graph walk to add context. They perform experiments on a dataset retrieved from Kialo, as we do in this work (see details in Section \ref{sec:kialo}). 

A similar task is rumour verification, where the goal is to evaluate the truthfulness of a rumour based on the reaction caused by it. 
In this case, since the focus is on the effects produced by the claim, the context is represented by the claims following the target claim (i.e., the right context), rather than the claims preceding it (i.e., the left context). 
To address this task, \citet{tian-etal-2022-duck} propose a combination of BERT
with a particular Graph Neural Network called GAT \citep{velivckovic2017graph}. They retrieve both linguistic and extra-linguistic context, but they consider the full discussion tree and perform classification only of the initial claim.

To summarize, existing past works that tried to integrate contextual information to classification tasks either were not able to outperform text-only approaches, or yielded an improvement using computationally expensive models such as Graph Neural Networks (GNNs). Furthermore, they tended to give in input to the model all possible information, including user data. With our approach, instead, \textit{context benefits classification}, while modelling the diverse types of input in \textit{natural language} and being \textit{privacy-preserving}.

\section{Problem statement}
\par The definition of \textit{discussion} is not unique. Depending on the social network, different \textit{discussion structures} can arise, from discussion chains to discussion trees, or allowing branches only at specific levels.
In the following, \textit{discussion chain} indicates a linear thread of ordered claims, where each claim is the reply to the previous one. This definition allows us to assume that the author of the $N^{th}$ claim has read all the previous $N-1$ claims.
Moreover, using the single chain instead of the discussion tree allows us to reduce the complexity of the discussion structure. From a discussion chain we can retrieve a Local Discussion Network (LDN), i.e. a multi-edge directed network of interaction among the users, with a timestamp label for each edge.
\vspace{0.5cm}
\par \textbf{Formalization.} Let $D = \{d_0, d_1, d_2, ..., d_m\}$ be a set of discussions, where each discussion is made of an ordered sequence of claims $d_i = \{\bar{c}_0, \bar{c}_1, \bar{c}_2, ..., \bar{c}_{n}\}$ where $\bar{c}_0$ is called \textit{initial claim} and each claim $\bar{c}_i$ is a response to the claim $\bar{c}_{i-1} \forall i \ge 1$. Each claim $\bar{c}_i$ is a tuple $\{c_i, u_i, t_i\}$, where $c_i$ is the textual content, $u_i$ the local user ID of the author and $t_i$ the timestamp. Each discussion $d_i$ has a label $y_i \in Y$, with $Y = [0, l-1]$ where $l$ is the number of possible labels. In Kialo setting (see details of Kialo dataset in Section \ref{sec:kialo}), we have two labels called \textit{contrast} (C) and \textit{support} (S) respectively mapped to $\{0, 1\}$. The goal is to learn a function $f$ that maps correctly each discussion to its correct label $f: D \rightarrow Y$.

\section{Kialo Dataset for Stance Detection}\label{sec:kialo}

\begin{figure}[t!]
\centering
\includegraphics[width=0.48\textwidth]{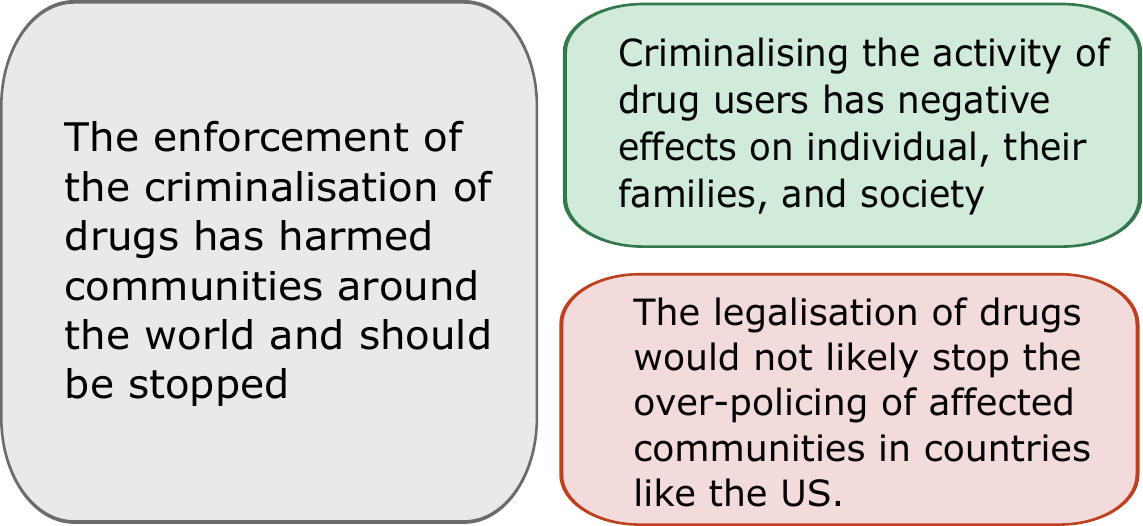}
\caption{Example of supportive (green) and contrastive (red) claim having the same parent claim in Kialo.}\label{fig:example_stance}
\end{figure}

Kialo\footnote{https://www.kialo.com} is an online platform where people can debate around a main topic, with moderators being in charge of checking the grammaticality of the claims, evaluating the level of support or of contrast between a target claim and its parent claim, and even moving claims to make conversations more consistent. 
For these reasons, Kialo typically contains less noisy data and a clearer conversational structure than other social media like Twitter/X, being an ideal testbed for experiments and analyses.  

In Kialo, the author of each comment is required to assign a \textit{stance label} to it with respect to the parent comment. This label (\textit{support} or \textit{contrast}) is then checked by the moderator, who can change it if needed (an example of supportive and contrastive stance from the dataset is displayed in Figure \ref{fig:example_stance}). 
Furthermore, being clearly structured, it is possible to easily retrieve from discussions the reply-tree structure and the distribution of \textit{support}/\textit{contrast} comments. 

Datasets extracted from Kialo have  already been used in the past to study the linguistic characteristics of impactful claims \cite{durmus-etal-2019-determining,durmus2019role} or perform polarity prediction  \cite{agarwal2022graphnli}. 
We obtained access to the dataset based on Kialo presented in \citet{scialom-etal-2020-toward}, which was used for binary stance detection. We extract from their data only a subset containing chains longer than $1$ (i.e., having at least the initial claim and one reply). In this way, we obtain  $122,681$ training instances, $7,447$ validation instances and $8,211$ test instances.  Each instance includes: \textsc{i.} the \textit{target} claim; \textsc{ii.} the \textit{discussion chain}, from the \textit{initial claim} to the \textit{target claim}; \textsc{iii.} the \textit{stance} of each claim versus its parent claim; \textsc{iv.} the \textit{user ID} of each claim; \textsc{v.} the \textit{timestamp} of each claim.
Given a discussion $d = \{\bar{c}_0, \bar{c}_1, ..., \bar{c}_{n}\}$ of length $n+1$, the goal is to classify correctly the stance of $\bar{c}_{n}$ with respect to $\bar{c}_{n-1}$, choosing between \textit{support} (S) or \textit{contrast} (C). 
We report in Table \ref{tab:label-distribution-kialo} an overview of the label  distribution in our dataset, which we call the Stance Detection Kialo dataset (SDK).

\begin{table}
\small
\centering
\begin{tabular}{| l | c | c | c |}
\hline
 & \multicolumn{3}{|c|}{\textbf{SDK Dataset}} \\
\hline
\textbf{Set}    & \textbf{Contrast}  & \textbf{Support}  & \textbf{Total} \\
\hline
Training        & $49.2$\%          & $50.8$\%         & $122,681$\\
Validation      & $50.2$\%        & $49.8$\%      & $7,447$\\
Test            & $54.5$\%        & $45.5$\%         & $8,211$\\
\hline
\end{tabular}
\caption{\label{tab:label-distribution-kialo}
Distribution of the labels in the Stance Detection Kialo (SDK) dataset. 
}
\end{table}

For each discussion tree we extract all the discussion chains going from the initial claim to the leaves. Consequently, it is possible for portions of these chains to overlap, while the target claims, with their respective labels, remain unique. This approach allows the model to process instances in which different discussion progressions result in different outcomes.
Furthermore, to mitigate potential data contamination effects, the dataset is split according to the initial claim $c_0$. As a result, all chains originating from the same initial claim are exclusively assigned to either training, validation, or test set.

\section{Context Definition and Modelling}\label{section:inputs}
 In past works, context has been integrated in social media classification tasks using two main approaches: by combining  linguistic and network information through the combination of node or network embeddings and textual embeddings \cite{shu2019beyond,dou2021user} or by using textual embeddings as features in a network system, and retrieving a general representation using GNNs or node/network embedding techniques \cite{Yao_Mao_Luo_2019,lin-etal-2021-bertgcn}.

We follow a third approach by expressing information on structural and temporal context using natural language, and then  giving it in input to a transformer-based model. We use a RoBERTa-based model \citep{liu2019roberta} to perform the task. This allows us to keep the same classification framework while only changing the input data to  progressively add contextual information, adopting a simple yet effective solution which is computationally lightweight.

Given a discussion chain $d = \{\bar{c}_0, \bar{c}_1, ..., \bar{c}_{n}\}$ of length $n+1$, where $\bar{c_i} = \{c_i, u_i, t_i\}$, we can identify $3$ different types of context: a linguistic (textual) context, $c_i$, and two extra-linguistic (temporal and structural) contexts, $t_i$ and $u_i$.

\par \textbf{Textual context.} In our experiments, the textual context is defined as the sequence of all the claims in the discussion chain from $c_0$ to $c_{n-2}$,
and it is added to $c_{n-1}$ and $c_n$ (i.e., the claims used for defining the stance). We concatenate all $c_i$ for $0\le i\le n$ and between each pair of claims we put a \verb|[SEP]| tag. If the length of the final input exceeds the maximum input length for the model, we iteratively delete $c_i$, for $i$ from $1$ to $n-2$ (keeping always $c_0$ at the beginning). We call this concatenation \textsc{txt\_chain}. 

\par \textbf{Temporal context.} To model the temporal context, we add at the beginning of each $c_i$ (from the textual context) the time $t_i$ passed between the publication of the initial claim $\bar{c}_0$ and of $\bar{c}_i$. However, we know that transformer-based models struggle in 
mathematical reasoning \cite{patel-etal-2021-nlp}. To overcome this limitation, instead of reporting $t_i$ as a value in milliseconds (as provided in the dataset) the temporal information is given in the format ``\texttt{after $d$ days, $h$ hours, $m$ minutes}'', with $d$, $h$, and $m$ correctly computed. 
We call this prefix \textsc{Time}. This prefix is delimited by two special tags: \verb|<t>| and \verb|</t>|.

\par \textbf{Structural context.} To model the structural context, we add at the beginning of each text $c_i$ the local user ID of $u_i$. This piece of information makes it possible to reconstruct the structure of the LDN among the users in the discussion $d$, i.e. if $A$ replies to $B$, there is a direct edge from $A$ to $B$. We can therefore see the LDN as a multi-edge directed graph of the interactions, with the textual content and the order of interactions as labels (Figure \ref{fig:enter-label}). 

The local user ID is \textit{locally unique}: 
for each discussion chain, a value from $0$ to $m-1$ is incrementally assigned to each of the $m$ users contributing in the discussion according to their first appearance within the discussion itself.
Using local IDs means that when a user is active across different discussions, they are assigned a different ID in each conversation. This prevents our model from implicitly profiling  users' behavior and attitude at global level, thus adopting a privacy-preserving approach. 

The structural information is given in input to the model adding before each comment the prefix  ``\texttt{$j$th user}'', with $0\le j \le m-1$ to declare that the author with local ID $j$ wrote the claim. We call this prefix \textsc{User}. Also for this prefix we adopt two special tags to signal the start and the end of the prefix: \verb|<o>| and \verb|</o>|. 

\section{Models and Experimental Settings}\label{experimental_setting}

\begin{figure}[t!]
    \centering
    \includegraphics[]{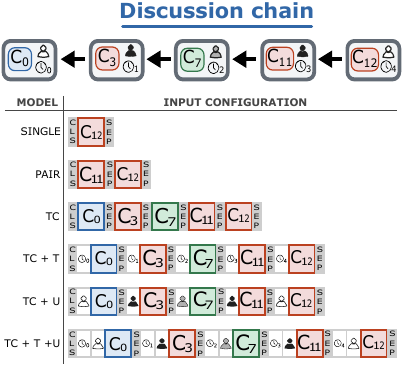}
    \caption{Schematic view of the input configuration for each model tested. We display the position of each textual content $c_i$, the \texttt{[CLS]} tokens, the \texttt{[SEP]} tokens, the \textsc{user} prefix and the \textsc{time} prefix.}
    \label{fig:input-configuration}
\end{figure}

\begin{table*}
    
\small
\begin{tabularx}{\textwidth}{| l | X |}
\hline
\textbf{Model} & \textbf{Input}\\
\hline
\textsc{Single} & \textbf{<s>}Receiving a benefit while helping others is not morally wrong. Otherwise all the foundations supported by big brands would be morally reprehensible. \textbf{</s>}
\\ 
\hline
\textsc{Pair} & \textbf{<s>} Personal "return on investment" should not be a guide on charity. \textbf{</s></s>} Receiving a benefit while helping others is not morally wrong. Otherwise all the foundations supported by big brands would be morally reprehensible. \textbf{</s>}\\
\hline
\textsc{TC} & \textbf{<s>} People should donate to organisations that support gorillas instead of to those that support starving children. \textbf{</s></s>} Saving gorillas has less impact on the donor's own well-being than saving a child. \textbf{</s></s>} Personal "return on investment" should not be a guide on charity. \textbf{</s></s>} Receiving a benefit while helping others is not morally wrong. Otherwise all the foundations supported by big brands would be morally reprehensible. \textbf{</s>}\\
\hline
\textsc{TC + T} & \textbf{<s> <t>} after 0 days, 0 hours, 0 minutes \textbf{</t>} People should donate to organisations that support gorillas instead of to those that support starving children. \textbf{</s></s> <t>} after 0 days, 6 hours, 36 minutes \textbf{</t>} Saving gorillas has less impact on the donor's own well-being than saving a child.  \textbf{</s></s> <t>} after 7 days, 20 hours, 23 minutes \textbf{</t>} Personal "return on investment" should not be a guide on charity.  \textbf{</s></s>} <t> after 28 days, 23 hours, 14 minutes \textbf{</t>} Receiving a benefit while helping others is not morally wrong. Otherwise all the foundations supported by big brands would be morally reprehensible. \textbf{</s>} \\
\hline
\textsc{TC + U} &  \textbf{<s> <o>} 0th user \textbf{</o>} People should donate to organisations that support gorillas instead of to those that support starving children. \textbf{</s></s> <o>} 1st user \textbf{</o>} Saving gorillas has less impact on the donor's own well-being than saving a child. \textbf{</s></s> <o>} 2nd user \textbf{</o>} Personal "return on investment" should not be a guide on charity.  \textbf{</s></s> <o>} 1st user \textbf{</o>} Receiving a benefit while helping others is not morally wrong. Otherwise all the foundations supported by big brands would be morally reprehensible. \textbf{</s>}\\
\hline
\textsc{TC + U + T} &  \textbf{<s> <t>} after 0 days, 0 hours, 0 minutes \textbf{</t> <o>} 0th user \textbf{</o>} People should donate to organisations that support gorillas instead of to those that support starving children. \textbf{</s></s> <t>} after 0 days, 6 hours, 36 minutes \textbf{</t> <o>} 1st user \textbf{</o>} Saving gorillas has less impact on the donor's own well-being than saving a child. \textbf{</s></s> <t>} after 7 days, 20 hours, 23 minutes \textbf{</t> <o>} 2nd user \textbf{</o>} Personal "return on investment" should not be a guide on charity. \textbf{</s></s> <t>} after 28 days, 23 hours, 14 minutes \textbf{</t> <o>} 1st user \textbf{</o>} Receiving a benefit while helping others is not morally wrong. Otherwise all the foundations supported by big brands would be morally reprehensible. \textbf{</s>}\\
\hline

\end{tabularx}
\caption{Different types of input related to the same discussion that are fed to the model.}\label{tab:input_examples}
\end{table*}

We implement and compare eight different classification models trained on the SDK dataset, which  can be divided into three categories: \textsc{Dummy}, \textsc{Baselines} and \textsc{Contextual}. \textsc{Dummy} models predict the label ignoring the input (i.e., majority class or random class). Instead, for \textsc{Baselines} and \textsc{Contextual} we always use a pre-trained RoBERTa-based model \citep{liu2019roberta} to embed the input. Then we extract the final \verb|[CLS]| contextual embedding and feed it into a Multi-Layer Perceptron (MLP) module to perform the classification task (for details of the architecture, see Appendix \ref{model_archit}). We use Optuna \citep{akiba2019optuna} for hyperparameter optimization of the learning rate and the dropout applied to the MLP (details in Appendix \ref{app:training_details}). In Figure \ref{fig:input-configuration} we report a schematic view of the input configuration employed for the \textsc{Baseline} models and the \textsc{Contextual} models. 

We describe below the different classification models, divided into the three following categories.

\par \textbf{\textsc{Dummy}.} We implement two ``dummy'' models:
\begin{itemize} 
    \item \textsc{Majority Class:} this model  always assigns the majority class label (i.e., \textit{support} in the case of the SDK dataset). 
    \item \textsc{Random:} this model assigns the label, for each item, at random, each with the probability $p=0.5$.
\end{itemize}
\par \textbf{\textsc{Text-Only Baselines}.} The two  models, based only on the text of the claims, take in input a fixed number of claims:
\begin{itemize} 
    \item \textsc{Single:} we give in input to the model only the textual content of the last claim $c_{n}$. The goal is to predict the stance of $c_{n}$ without considering what was written before. This approach should be able to perform classification just by looking at linguistic or stylistic cues in $c_{n}$.

    \item \textsc{Pair:} we give in input to the model only the textual content of the last two comments, $c_{n}$ and $c_{n-1}$, separated by the \verb|[SEP]| token. The goal here is to predict the correct label looking at the semantics and at the style of the two claims, as well as at the relations between the two. 
    This is the standard  solution for Stance Detection. 
\end{itemize}

\par \textsc{\textbf{Contextual.}} We model contextual information in four different ways:
\begin{itemize}
\itemsep0em 
    \item \textsc{TC:} we give in input to the model only the concatenated claims in the \textsc{txt\_chain} format.
    \item \textsc{TC + T:} we give in input to the model the concatenated claims in the \textsc{txt\_chain} format, each claim with the \textsc{time} prefix.
    \item \textsc{TC + U:} we give in input to the model the concatenated claims in the \textsc{txt\_chain} format, each claim with the \textsc{user} prefix.
    \item \textsc{TC + U + T:} we give in input to the model the concatenated claims in the \textsc{txt\_chain} format, each claim with the \textsc{time} prefix and the \textsc{user} prefix.
\end{itemize}

We report in Table \ref{tab:input_examples} an example of how the same discussion is given in input to the model in the different configurations. In the pretrained RoBERTa model available on Hugging Face\footnote{\url{https://huggingface.co/docs/transformers/model\_doc/roberta}}, the \texttt{[CLS]} token is replaced by a \texttt{<s>} tag and the \texttt{[SEP]} token is represented by a sequence of special tags (i.e., \texttt{</s></s>}). We have taken inspiration from these representations for our new special tokens: \texttt{<t>}, \texttt{</t>}, \texttt{<o>}, \texttt{</o>}. The input text is pre-processed by replacing user mentions and urls with placeholders following a standard approach for social media data.\footnote{\url{https://huggingface.co/cardiffnlp/twitter-roberta-base-sentiment}}

\section{Experiments}\label{sec:first_experiment}
\subsection{Stance Detection on Kialo}

\begin{table*}[t!]
\small
\centering
\begin{tabular}{| l | l | l | l | l | 
l | c | c |}
\hline
\textbf{Category}   &\textbf{Model} &\textbf{C-F1}  &\textbf{S-F1}  &\textbf{W-F1}  &\textbf{M-F1}  &\textbf{LR}    &\textbf{DO}\\
\hline
\multirow{2}{*}{\textsc{Dummy}} &\textsc{Majority}    &$70.5\;(\pm0.0)$   &$0.0\;(\pm0.0)$    &$38.4\;(\pm0.0)$  &\cellcolor{col_table!10}$35.3\;(\pm0.0)$ &$/$    &$/$\\
\cline{2-8}
& \textsc{Random} & $52.1\;(\pm0.6)$ & $48.0\;(\pm0.4)$ & $50.2\;(\pm0.5)$ & \cellcolor{col_table!30}$50.1\;(\pm0.5)$ & $/$& $/$\\
\hline
\multirow{2}{*}{\textsc{Baselines}}
& \textsc{Single} & $75.5\;(\pm0.5)$ & $70.2\;(\pm0.6)$ & $73.0\;(\pm0.1)$ &\cellcolor{col_table!50} $72.8\;(\pm0.2)$ &$7.5\cdot 10^{-6}$ & $0.5$\\
\cline{2-8}
& \textsc{Pair} & $83.1\;(\pm0.4)$ & $79.3\;(\pm0.4)$ & $81.4\;(\pm0.2)$ &\cellcolor{col_table!70} $81.2\;(\pm0.2)$ & $7.5\cdot 10^{-6}$ & $0.25$\\
\hline
\multirow{4}{*}{\textsc{Contextual}} & \textsc{TC} & $82.2\;(\pm0.6)$ & $78.8\;(\pm0.4)$ & $80.7\;(\pm0.3)$ &\cellcolor{col_table!65} $80.5\;(\pm0.3)$ & $7.5\cdot 10^{-6}$ & $0.25$\\
\cline{2-8}
& \textsc{TC + T} & $83.3\;(\pm0.4)$ & $80.0\;(\pm0.4)$ & $81.8\;(\pm0.3)$ &\cellcolor{col_table!75} $81.7\;(\pm0.3)^{\ast }$ & $7.5\cdot 10^{-6}$ & $0.25$\\
\cline{2-8}
& \textsc{TC + U} & $85.2\;(\pm0.5)$ & $82.1\;(\pm0.7)$ & $83.8\;(\pm0.5)$ &\cellcolor{col_table!90} $83.7\;(\pm0.5)^{\ast \diamond}$ & $1.0\cdot 10^{-5}$ & $0.25$\\
\cline{2-8}
& \textsc{TC + U + T} & $85.6\;(\pm0.4)$ & $82.3\;(\pm0.3)$ & $84.0\;(\pm0.3)$ & \cellcolor{col_table!92} $83.9\;(\pm0.3)^{\ast \diamond}$& $7.5\cdot 10^{-6}$ & $0.25$\\
\hline

\end{tabular}
\caption{F1 scores obtained on the test set of SDK dataset, for each class, in weighted average and in macro average (average of the best 5 runs in validation over 10). ($\ast$) and ($\diamond$) show a statistically significant improvement with respect to the \textsc{Pair} baseline, for ASO test and Student's t-test respectively. We report the average and the standard deviation for each metric. LR column reports the Learning Rate and DO column reports the dropout value in the MLP component}\label{tab:test_results_sdk}

\end{table*}

The goal of the first set of experiments is to evaluate on Kialo the performance of the eight models described above by using the whole training set, both for hyperparameter optimization and for the final evaluation.
The results are the average and standard deviation over $5$ experimental runs (details in Appendix \ref{app:training_details}).
We report in Table \ref{tab:test_results_sdk} the F1 score for each class,
its weighted average (W-F1), and the macro average (M-F1). The final metric we use for ranking the models is M-F1.

\textbf{Results.} 
All the results are reported in Table \ref{tab:test_results_sdk}. We compute statistical significance using Almost Stochastic Order test \cite{del2018optimal, dror-etal-2019-deep} and Student's t-test for independent sample with Bonferroni correction \cite{1570009749360424576}. For ASO, we use the implementation provided in the \texttt{deep-significance} library, presented by \citet{42042c11faa84a09a9b1e9a759cdf51d}, with the suggested threshold value of $\tau = 0.2$. For the t-test we use the implementation provided in the \texttt{scipy} library with threshold value of $\alpha = 0.05$.

Both \textsc{Baseline} models lead to better performances than the  \textsc{Dummy} models. Interestingly, the \textsc{Single} model performs well ($72.8$ M-F1 on average), showing that the style of the target comment already conveys relevant information to detect its stance. However, as expected, taking the last two comments in input (\textsc{Pair} model) increases the M-F1 score by $+8.4$ over the \textsc{Single} one. 

Among the \textsc{Contextual} models, the \textsc{TC} model achieves the worst results, slightly lower than the \textsc{Pair} model. This shows that adding context is not always beneficial. In this case, since the number of claims in a discussion changes, the model is probably not able to focus on the right portion of the chain. 
%Similar results were observed also in  \citet{menini2021abuse}.  
Adding the temporal information only, as in the \textsc{TC + T} model, yields a better performance than the simple textual chain in the \textsc{TC} model ($+1.2$ M-F1) and outperforms significantly the \textsc{Pair} baseline ($+0.5$) for the ASO test.

Looking at the different types of context, we observe that adding only the \textsc{user} prefix as in \textsc{TC + U}, leads to a significant increase of $+3.2$ M-F1 over the \textsc{TC} model and of $+2.5$ over the \textsc{Pair} baseline, for both  statistical significance tests. Furthermore, the \textsc{TC + U + T} model with both \textsc{user} prefix and \textsc{time} prefix increases significantly the performance with respect to \textsc{TC} model ($+3.4$), \textsc{Pair} model ($+2.7$) and \textsc{TC + T} model ($+2.2$), again for both statistical significance tests. However, there is no significant difference between \textsc{TC + U} model and \textsc{TC + U + T} model (only $+0.2$). This indicates that \textsc{time} prefix is no more relevant once we pass to the model the \textsc{user} prefix.

\subsection{Experiments on other Datasets}\label{sec:otherdatasets}

As a comparison, we run the same experiments on two smaller datasets, which provide the same type of information included in SDK: the SQDC dataset \cite{gorrell-etal-2019-semeval} for stance detection, and the ContextAbuse dataset  \cite{menini2021abuse} for abusive language  detection. These datasets present a size of respectively 5\% and 7\% compared to SDK. 
On the SQDC dataset, the \textsc{Single} baseline yields the best result (47.2 M-F1), probably because the official test set contains only chains of length 2. After creating a better balanced train and test split, instead, the best result is obtained with the  \textsc{Pair} baseline (46.4 M-F1). On the ContextAbuse dataset, adding textual context (i.e., \textsc{TC} model) yields the best performance (81.4 M-F1), which however is not statistically significant compared to the \textsc{Single} baseline (80.7 M-F1). For detailed dataset specifications and experimental results, we refer to Appendix \ref{app:rumoureval} and Appendix \ref{app:contextabuse}.

These experiments suggest that, independently from the specific task, contextual information may not yield substantial enhancements in performance if the amount of training data is too limited. In order to investigate better this aspect, we perform an additional analysis of the learning curve in the following section.

\section{Learning Curve Analysis} \label{sec:second_experiment}

\begin{figure}
    \centering
    \includegraphics[width = 0.5\textwidth]{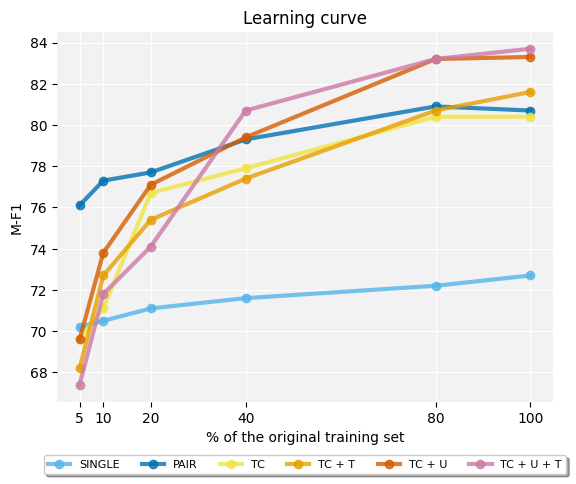}

    \caption{Learning curve for each \textsc{Baseline} and \textsc{Contextual} model, in terms of M-F1 score.}\label{fig:learning_curve_sdk}
\end{figure}

While our experiments show that the discussion context on the SDK dataset is beneficial to stance detection, we aim to assess the impact of the training set size. Our intuition is that, when contextual information is embedded in the model, more training instances are needed than for text-only models.  Indeed, the model must be given enough training instances to understand what is the role of the special tags and what type of information is included between two specific separators. 

We therefore extract from the original training data $5$ different training sets, comprising around $5\%$ ($6,354$ examples), $10\%$ ($12,402$ examples), $20\%$ ($24,748$ examples), $40\%$ ($49,249$ examples) and $80\%$ ($98,389$ examples) of the original training instances. 

\begin{figure*}[h!]
    \centering
    \includegraphics[width = \textwidth]{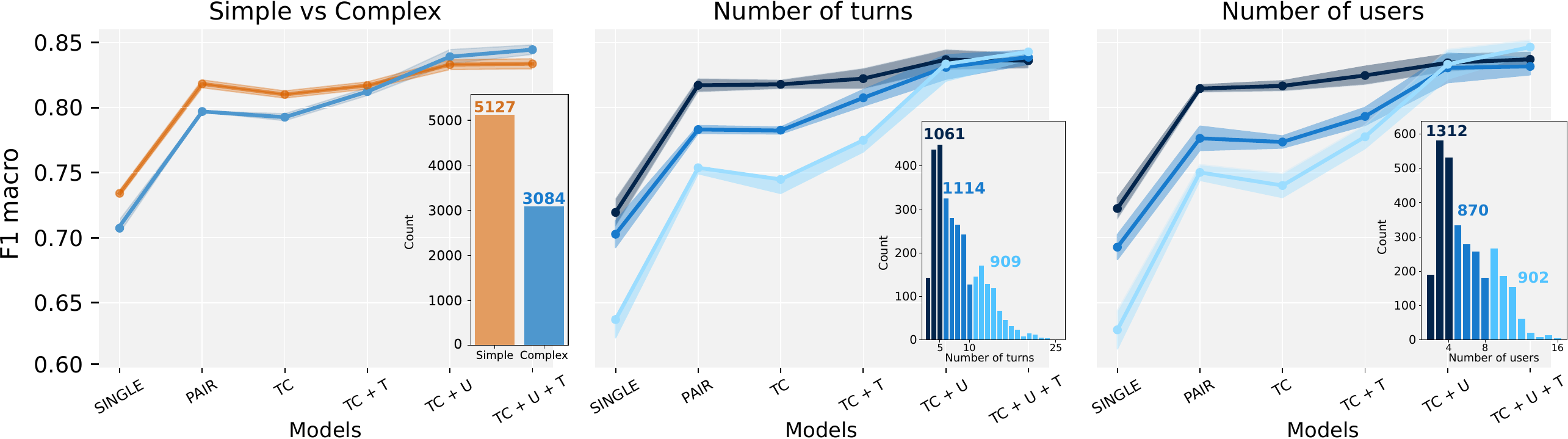}
    \caption{Model comparison when testing the classifier on different dimensions: Simple vs. Complex LDNs (left), Complex LDNs with different number of turns (center) and  different number of users (right).}
    \label{fig:conv_struct}
\end{figure*}

\textbf{Results.} 
Figure \ref{fig:learning_curve_sdk} shows the results obtained when increasing the training set size 
 as the average over 3 runs (the full results and experimental details are reported in Appendix \ref{app:learning_curve}). We exclude the \textsc{Dummy} models, since they never outperform \textsc{Baseline} and \textsc{Contextual} models.

With $5\%$ of the training data, all the \textsc{Contextual} models are beaten by the worst \textsc{Baseline} model (i.e., \textsc{Single}), with a drop in M-F1 ranging from $-10.8$ (TC) to $-16.3$ (TC+U+T) compared to using the whole  training set. At the same time, the \textsc{Pair} model achieves the best result in this setting, with a performance drop of only $-4.6$. However, as soon as we add more data, the scenario changes. With $10\%$ training set and $20\%$ training set, \textsc{Contextual} models overcome the \textsc{Single} model and progressively approach the \textsc{Pair} model. With $40\%$ training set, \textsc{TU + U} and \textsc{TC + U + T} outperform the \textsc{Pair} model and with more data they substantially increase their gap with the latter. 
%Only \textsc{TC + T} outperforms the \textsc{Pair} model with $100\%$ of the training data.
\\ \indent To sum up, these results show that \textsc{Contextual} models need between $20\%$ and $40\%$ of the training data (i.e., from 24 thousand to 49 thousand training examples) to achieve comparable results with the \textsc{Pair} model, while they need more data to outperform it.
%\footnote{

%} 

\section{Analysis of Discussion Structure }
\label{sec:last_experiment} 

Beside assessing the impact of training set size on classification performance, we are also interested in analysing the role played by the topology of local discussion networks (LDNs) in terms of complexity and discussion length. To this aim, we merge consecutive claims written by the same author in a discussion chain into a unique \textit{turn}, and create a corresponding turn chain. In this way, two consecutive turns have always different authors, and the corresponding LDN does not have self-loops. For further details we refer to Appendix \ref{app:analysis}.

We first divide LDNs in the SDK dataset into two groups: simple LDNs, which are characterized by chains where users write only one turn, and complex LDNs, with a user writing  several turns.
We run the stance detection experiment with the setting presented in Section \ref{sec:first_experiment} and compare the results obtained on simple vs. complex chains. We also analyse how the number of claims and of users affects classifier performance on complex LDNs (with and without context).  
Results are reported in Figure \ref{fig:conv_struct}, which displays the M-F1 score obtained with the different models. The thickness of the line represents the standard deviation over $5$ runs. 
The analysis shows that extra-linguistic context gives an important contribution to the classification of complex LDNs, in particular the \textsc{TC + U + T} model. This contribution is more limited on simple chains, with the \textsc{Pair} model and the \textsc{Contextual} models achieving comparable results.

As regards the impact that the number of turns  has on the classification of complex LDNs  (middle graph in Figure \ref{fig:conv_struct}), we first group the turns into three bins based on their length: from $2$ to $5$ (dark blue), from $6$ to $10$ (blue) and > $10$ (light blue). 
The comparison among the three groups clearly demonstrates that the inclusion of  temporal and structural context consistently results in a performance improvement, regardless of the number of turns in the discussion. 
We finally investigate the effect that the number of users involved in the complex LDN has on classification performance (right plot of Figure \ref{fig:conv_struct}). Also in this case, the chains are grouped into three bins: having less that $4$ users (dark blue), from $5$ to $8$ users (blue), and more than $8$ (light blue). Again, the comparison demonstrates that the inclusion of the extra-linguistic contexts consistently results in improvement, regardless of the number of users involved in the discussion.

\section{Discussion}

The results reported in Section \ref{sec:first_experiment} and Section \ref{sec:second_experiment} show that adding extra-linguistic context is beneficial to improve performance on stance detection. However, this benefit arises only if the \textsc{Contextual} models have access to enough data, which in our experiments on the SDK dataset means between $24,000$ and $49,000$ items. This result explains also the different performance obtained on smaller datasets (Section \ref{sec:otherdatasets}).
As regards the analysis of local discussion chains, the more complex is the LDN, the more evident are the benefits from the structural context. This suggests that our transformer-based model is able to capture the structure given by the interactions among the users, even if implicit, when enough data are available. 
Our analyses show also that capturing contextual information is particularly beneficial with longer chains of turns, and discussion chains with more users. When all contextual information (both linguistic and extra-linguistic) is included in the model, the classifier performs equally well on long and on short chains, making the results more consistent and the model more robust to chain length and user activity. 

As regards the temporal context, we show that it is still useful to achieve a better performance, but we argue that in Kialo it may not be particularly relevant because this is a platform where users are more likely to ponder their responses and take some time to reflect before posting, also thanks to a strict moderation policy  \cite{Vosoughi2018TheSO}.

\section{Conclusions}
In this paper we have tested the effectiveness of using linguistic and extra-linguistic contexts for text classification.
Our results show that full linguistic context alone worsens or does not significantly improve the results with respect to the non-contextual baseline.
Instead, with extra-linguistic context, the performance improves, especially with the contribution of structural context. Further analysis shows that such results strongly depend on the amount of data on which the models are trained. Moreover, we found that extra-linguistic context makes results more robust across discussion networks of different lengths and more or less active users.
Our experiments show also that transformer-based models are able to embed structural features, which can be effectively given in input to the model in the form of simple natural language statements.

\section{Limitations}

The findings presented in this work were mainly focused on the Kialo dataset on the specific task of stance detection.
Kialo is an ideal testbed for our hypotheses because it is a moderated platform with well-structured discussions written in plain English. It is not possible to infer that the same findings would be confirmed on any social network, where discussions may be more fragmented and lacking moderation. Indeed, to have a clear picture of our findings, 
other large datasets with similar characteristics would be needed. Nevertheless, as a preliminary exploration, our experiments on the two smaller datasets from Twitter/X confirmed our expectation about the importance of the amount of training data.
Moreover, our work presents a limited number of classification models. We tested a few other  combinations without reaching interesting results, therefore we decided to focus only on few configurations and to analyse their behavior more thoroughly. Overall, our contribution is not focused on generally achieving the best results, but rather on assessing how and why contextual information influences the behavior of a  model.

\section{Ethics Statement}

Integrating user information into a text classification task may pose ethical risks, since profiling may introduce biases in classification, hurting some individuals with a specific profile, and is explicitly prohibited in a number of countries. However, we adopt a solution that minimises such risks in that it does not use global user information but only local one, making it impossible to infer user information at platform level. Furthermore, no additional information about users' preferences and attitude is explicitly coded: the model is given in input only \textit{what} and \textit{when} users post in each discussion, and in response \textit{to whom}.   

 In terms of reproducibility, our models are extremely lightweight and allow the reproduction of the experiments on common GPUs, using implementations available online.

\section*{Acknowledgements}
We thank the reviewers for their insightful suggestions during the ARR process.
We acknowledge the support of the PNRR project FAIR - Future AI Research (PE00000013), under the NRRP MUR program funded by the NextGenerationEU.
This work was also funded by the European Union’s Horizon Europe research and innovation program under grant agreement No. 101070190 (AI4Trust) and under grant agreement No.  101120237 (ELIAS).
Nicolò Penzo's activities are part of the network of excellence of the European Laboratory for Learning and Intelligent Systems (ELLIS).

\bibliography{anthology,custom}

\clearpage

\appendix

\section{Appendix}
\label{sec:appendix}

\subsection{Model Architecture}\label{model_archit}

\begin{figure}
    \centering
    \includegraphics[width = 0.47\textwidth]{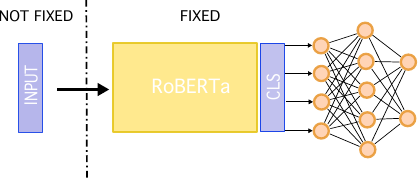}
    \caption{Schematic view of the model we tested. We distinguish between the component we change in each experiment (the input) and the fixed structure (RoBERTa + MLP).}
    \label{fig:model_architecture}
\end{figure}

The model architecture is reported schematically in Figure \ref{fig:model_architecture}. It is made of two main components: a RoBERTa model with on top a Multi Layer Perceptron (MLP). To perform the prediction, we feed the RoBERTa model with the input, and then we extract the final \verb|[CLS]| contextual embedding. So we pass the \verb|[CLS]| contextual embedding to the MLP, which consists in a classic Feedforward Neural Network (FNN), and perform the prediction.
\par The dimension of the \verb|[CLS]| contextual embedding is $d=768$. The RoBERTa model architecture and initial weights correspond to the pretrained version provided by Hugging Face called \verb|roberta-base|\footnote{\url{https://huggingface.co/roberta-base}}, with maximum input length $l=512$ tokens.
\par The MLP consists in $3$ layers: \textsc{i.} the first goes from dimension $768$ to $200$ with \verb|ReLU| activation function; \textsc{ii.} the second goes from dimension $200$ to dimension $300$, again with \verb|ReLU| activation function; \textsc{iii.} the third goes from dimension $300$ to dimension $n$, where $n$ is the number of classes among which we predict the class, with \verb|tanh| activation function. Finally we apply a \verb|softmax| on the $n$ value in output from the last layer, in order to have a probability distribution among the $n$ possible values (the prediction will correspond to the index of highest probability).

\subsection{Training Details.} \label{app:training_details}
\textbf{Hyperparameter search and Evaluation.} 
We exploit Optuna \citep{akiba2019optuna} for hyperparameter search, using a grid search for: \textsc{i.} the learning rate, with a uniform probability between the values $7.5\cdot 10^{-6}$, $1.0\cdot 10^{-5}$, $2.5\cdot 10^{-5}$, $5\cdot 10^{-5}$, $7.5\cdot 10^{-5}$; \textsc{ii.} the dropout applied between the layers of the MLP, with values $0.25$ and $0.5$. We use batch size $b = 32$ and weight decay $w_d = 10^{-4}$ in the RoBERTa components. In SDK dataset, we use unweighted Cross Entropy loss both in the training and in the validation phase, since the imbalance is negligible. 
\par For the final evaluation, we fix the hyperparameters and then we perform $10$ runs, changing each time the random seed. Then we keep the 5 best runs in validation, in order to exclude possible ``outlier" runs due to initialization problems. We compute the average and standard deviation of the test results on these 5 best runs.

\textbf{Training pipeline.} We perform backpropagation on the full structure of the model, without freezing any layer. As previously stated, our experiments keep always the same model, just changing the input. We use early stopping for model selection with patience $p=2$ epochs for the SDK dataset (Section \ref{sec:first_experiment} and Section \ref{sec:second_experiment}) and $p=5$ epochs for the SQDC dataset (Appendix \ref{app:rumoureval}). In the SDK dataset, each epoch corresponds to a training epoch on a sample of the training set, which is around half of the total training set, in order to speed up computation and generalization. We test also the usage of the full training set in each epoch, but the results remain comparable. This holds for all the experiments on Kialo datasets, the standard one (Section \ref{sec:first_experiment}) and the learning curve on training size (Section \ref{sec:second_experiment}). For the SQDC dataset and ContextAbuse dataset, we refer respectively to Appendix \ref{app:rumoureval} and Appendix \ref{app:contextabuse}.

\par For all the experiments we use a single A40 GPU with 48GB Memory. All the experimental code is developed in PyTorch. It requires around $33$ minutes of computation for each epoch (training phase plus validation phase).

\subsection{Analysis of truncation effects on SDK dataset}\label{app:analysis_truncation}

\begin{figure*}[t!]
    \includegraphics[width=\textwidth]{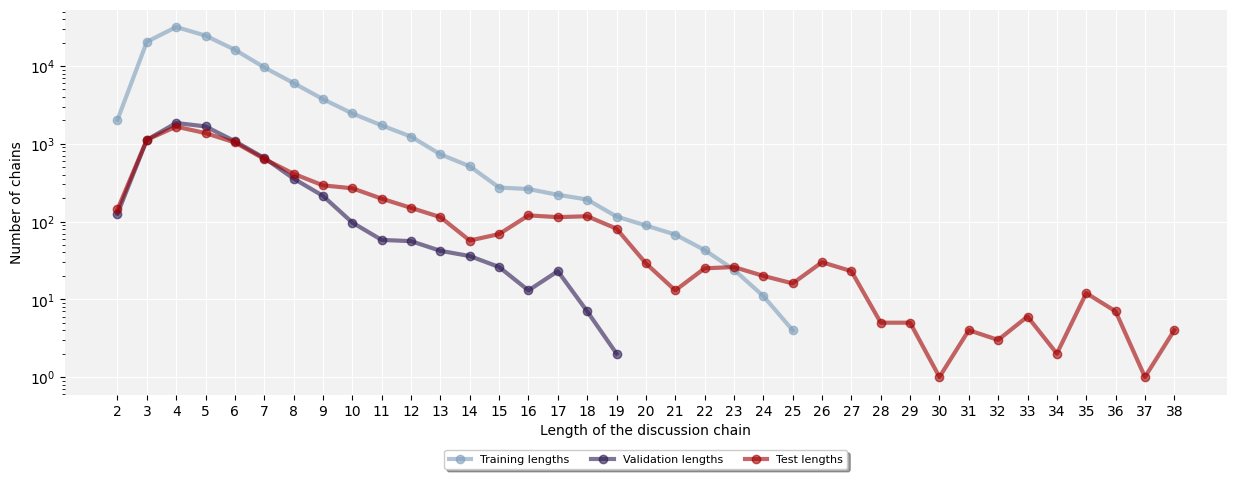}\hfill
    \caption{Length distribution of the discussion chains (i.e. number of claims in the discussion chain) in SDK dataset.}\label{fig:length_distribution}
\end{figure*}

In Section \ref{section:inputs}, we discuss the processing of strings that exceed the maximum input length by employing a deterministic truncation process on the discussion chains until the length satisfies the model constraint. We conduct an additional evaluation to investigate whether such truncation correlates with the final results, implying potential effects on performance.

For each contextual input configuration and dataset split, we compute the following metrics:
\textsc{i.} Truncation rate (ratio of truncated sequences); \textsc{ii.} average truncation (number of truncated claims); \textsc{iii.} average original length of the truncated sequences.

The statistics in Table \ref{tab:truncation-averages-subtables} reveal that the \textsc{TC + T} and \textsc{TC + U + T} input configurations result in more truncated chains, while \textsc{TC + U} exhibits less truncation than \textsc{TC + T}. Nevertheless, both \textsc{TC + U} and \textsc{TC + U + T} configurations perform similarly and outperform the \textsc{TC + T} model. This analysis suggests that the impact of the truncation process does not significantly influence our findings.

We report in Figure \ref{fig:length_distribution} the plot of the original lengths of the discussion chains (in terms of number of claims).

\begin{table}[h]
    \small
    \centering
    \begin{subtable}
        \centering
        \begin{tabular} {|p{2.2cm}|p{0.8cm}|p{0.9cm}|p{0.9cm}|}
            \hline
            %\toprule
            \textsc{TC} & \textbf{Train} & \textbf{Valid.} & \textbf{Test} \\
            \hline
            %\midrule
            \textbf{Truncation Rate} & $1.01\%$ & $0.70\%$ & $8.30\%$ \\
            \hline
            \textbf{Avg Truncation} & $4.20$ & $4.29$ & $6.27$ \\
            \hline
            \textbf{Avg Original} & $16.68$ & $13.63$ & $19.98$ \\
            \hline
            %\bottomrule
        \end{tabular}
    \end{subtable}
    
    \begin{subtable}
        \centering
        \begin{tabular}{|p{2.2cm}|p{0.8cm}|p{0.9cm}|p{0.9cm}|}
            \hline
            %\toprule
            \textsc{TC + T} & \textbf{Train} & \textbf{Valid.} & \textbf{Test} \\
            \hline          
            %\midrule
            \textbf{Truncation Rate} & $3.60\%$ & $3.60\%$ & $13.68\%$ \\
            \hline
            \textbf{Avg Truncation} & $4.05$ & $3.30$ & $7.38$ \\
            \hline
            \textbf{Avg Original} & $13.48$ & $12.68$ & $17.06$ \\
            \hline
            %\bottomrule
        \end{tabular}
    \end{subtable}
    
    \begin{subtable}
        \centering
        \begin{tabular}{|p{2.2cm}|p{0.8cm}|p{0.9cm}|p{0.9cm}|}
            \hline
            %\toprule
            \textsc{TC + U} & \textbf{Train} & \textbf{Valid.} & \textbf{Test} \\
            \hline
            %\midrule
            \textbf{Truncation Rate} & $2.33\%$ & $1.92\%$ & $11.40\%$ \\
            \hline
            \textbf{Avg Truncation} & $4.14$ & $3.68$ & $7.20$ \\
            \hline
            \textbf{Avg Original} & $14.65$ & $13.89$ & $18.19$ \\
            \hline
            %\bottomrule
        \end{tabular}
    \end{subtable}
    
    \begin{subtable}
        \centering
        \begin{tabular}{|p{2.2cm}|p{0.8cm}|p{0.9cm}|p{0.9cm}|}
            \hline
            %\toprule
            \textsc{TC + U + T} & \textbf{Train} & \textbf{Valid.} & \textbf{Test} \\
            \hline
            %\midrule
            \textbf{Truncation Rate} & $6.70\%$ & $6.03\%$ & $18.40\%$ \\
            \hline
            \textbf{Avg Truncation} & $3.74$ & $3.44$ & $7.00$ \\
            \hline
            \textbf{Avg Original} & $11.95$ & $11.54$ & $15.31$ \\
            \hline
            %\bottomrule
        \end{tabular}
    \end{subtable}
    \caption{Statistics of the truncation process in the SDK dataset, with a separate table dedicated to each model and a column corresponding to each dataset split.}
    \label{tab:truncation-averages-subtables}
\end{table}

\begin{table*}[ht]
\small
\centering
\begin{tabular}{| l | l | l | l | l | 
l | l | l |}
\hline
\textbf{Category} & \textbf{Model} & $\textbf{5}\%$ & $\textbf{10\%}$ & $\textbf{20\%}$ & $\textbf{40\%}$ & $\textbf{80\%}$ & $\textbf{100\%}$\\
\hline
\multirow{2}{*}{\textsc{Dummy}}
& \textsc{Majority} & $35.3$ & $35.3$ & $35.3$ & $35.3$ & $35.3$ & $35.3$\\
\cline{2-8}
& \textsc{Random} & $50.1$ & $50.1$ & $50.1$ & $50.1$ & $50.1$ & $50.1$\\
\hline
\multirow{2}{*}{\textsc{Baselines}}
& \textsc{Single} & $70.2$ & $70.5$ & $71.1$ & $71.6$ & $72.2$ & $72.7$ \\
\cline{2-8}
& \textsc{Pair} & $76.1$ & $77.3$ & $77.7$ & $79.3$ & $80.9$ & $80.7$ \\
\hline
\multirow{4}{*}{\textsc{Contextual}} & \textsc{TC} & $69.6$ & $71.1$ & $76.7$ & $77.9$ & $80.4$ & $80.4$ \\
\cline{2-8}
& \textsc{TC + T} & $68.2$ & $72.7$ & $75.4$ & $77.4$ & $80.7$ & $81.6$ \\
\cline{2-8}
& \textsc{TC + U} & $69.6$ & $73.8$ & $77.1$ & $79.4$ & $83.2$ & $83.3$ \\
\cline{2-8}
&  \textsc{TC + U + T} & $67.4$ & $71.8$ & $74.1$ & $80.7$ & $83.2$ & $83.7$\\
\hline
\hline
\multicolumn{2}{|l|}{\textsc{Training Set Size}} & $6354$ & $12402$ & $24748$ & $49249$ & $98389$ & $122681$\\
\hline

\end{tabular}
\caption{Macro-F1 scores obtained on the test set of SDK dataset, for every training set in growing size.}\label{tab:test_results_learning_curve_sdk}

\end{table*}

\subsection{Learning curve experiment}\label{app:learning_curve}
We report in Table \ref{tab:test_results_learning_curve_sdk} the detailed results from the second experiment on the SDK dataset presented in Section \ref{sec:second_experiment}. We first run hyperparameter optimization on each training set. Then, after fixing the hyperparameters as in Section \ref{experimental_setting}, we perform $3$ experimental runs on each training set, changing the random seed each time, and compute the average M-F1 among the $3$ runs. The same evaluation is performed using the complete training set.

\subsection{Details about the analysis of the results on SDK dataset}\label{app:analysis}
In Kialo, the same author can write several consecutive comments, even in contrast with  each other. However, we are more interested in interactions among different users. For this reason, we introduce the concept of \textit{turn}. Given a discussion chain of $n$ claims, we can retrieve a chain of $n'$ turns, where two consecutive turns have different authors. This is possible by merging all consecutive claims written by the same user into a unique turn. For instance if we have a discussion chain $d$ of length $6$ with user sequence $\{u_0, u_0, u_1, u_1, u_1, u_2\}$, the associated turn chain has length $3$ merging into one turn the first two claims, then the following three into another turn and the last one is already a turn, with user sequence $\{u_0, u_1, u_2\}$. This represents also a simple discussion. A complex discussion might be similar to the following: if the user sequence is $\{u_0, u_1, u_0, u_0, u_2, u_2\}$, in the turn chain the user sequence becomes $\{u_0, u_1, u_0, u_2\}$.

\begin{table*}[!ht]
\tiny
\centering
\begin{tabular}{| l | l | l | l | l | 
l | l | l | c | c |}
\hline
\textbf{Category} & \textbf{Model} & \textbf{S-F1} & \textbf{Q-F1} & \textbf{D-F1} & \textbf{C-F1} & \textbf{W-F1} & \textbf{M-F1} & \textbf{LR} & \textbf{DO}\\
\hline
\multirow{2}{*}{\textsc{Dummy}}
& \textsc{Maj.} & $0.0\;(\pm0.0)$ & $0.0\;(\pm0.0)$ & $0.0\;(\pm0.0)$ & $86.6\;(\pm0.0)$ & $66.1\;(\pm0.0)$ & \cellcolor{col_table!20}$21.6\;(\pm0.0)$ & $/$& $/$\\
\cline{2-10}
& \textsc{Rand.} & $12.6\;(\pm2.1)$ & $9.5\;(\pm1.1)$ & $13.9\;(\pm2.6)$ & $37.5\;(\pm1.9)$ & $31.6\;(\pm1.3)$ & \cellcolor{col_table!18}$18.3\;(\pm0.6)$ & $/$& $/$\\
\hline
\multirow{2}{*}{\textsc{Basel.}}
& \textsc{Single} & $14.1\;(\pm7.7)$ & $54.4\;(\pm2.9)$ & $47.5\;(\pm3.5)$ & $72.6\;(\pm5.7)$& $64.1\;(\pm4.2)$ & \cellcolor{col_table!60}$47.2\;(\pm2.3)$ & $5.0\cdot 10^{-5}$ & $0.25$ \\
\cline{2-10}
& \textsc{Pair} & $13.5\;(\pm1.6)$ & $58.4\;(\pm3)$ & $44.9\;(\pm0.1.5)$ & $71.1\;(\pm3.2)$ & $62.8\;(\pm2.3)$ & \cellcolor{col_table!57}$47.0\;(\pm0.5)$ &  $2.5\cdot 10^{-5}$ & $0.25$ \\
\hline
\multirow{4}{*}{\textsc{Cont.}} & \textsc{TC} & $12.9\;(\pm4.1)$ & $58.6\;(\pm2.4)$ & $42.7\;(\pm7.2)$ & $71.5\;(\pm4.3)$ & $62.9\;(\pm4.2)$ & \cellcolor{col_table!50}$46.4\;(\pm4.0)$ & $1.0\cdot 10^{-5}$ & $0.25$\\
\cline{2-10}
& \textsc{TC + T} & $15.4\;(\pm0.8)$ & $59.0\;(\pm2.6)$ & $44.1\;(\pm4.5)$ & $63.4\;(\pm3.7)$ & $57.0\;(\pm2.8)$ & \cellcolor{col_table!45}$45.5\;(\pm1.6)$ & $1.0\cdot 10^{-5}$ & $0.5$ \\
\cline{2-10}
& \textsc{TC + U} & $13.2\;(\pm5.1)$ & $56.3\;(\pm4.6)$ & $41.6\;(\pm3.8)$ & $65.1\;(\pm11.4)$ & $57.8\;(\pm8.6)$ & \cellcolor{col_table!38}$44.0\;(\pm3.3)$ & $2.5\cdot 10^{-5}$ & $0.25$ \\
\cline{2-10}
& \textsc{TC + U + T} & $19.2\;(\pm4.7)$ & $52.3\;(\pm3.5)$ & $43.1\;(\pm1.8)$ & $68.6\;(\pm4.7)$ & $61.1\;(\pm3.9)$ & \cellcolor{col_table!47}$45.8\;(\pm2.3)$ & $2.5\cdot 10^{-5}$ & $0.5$ \\
\hline

\end{tabular}
\caption{\textbf{SQDC - Challenge.} F1 scores obtained on the test set of SQDC dataset, on the original split given for the challenge. The F1 score is reported for each class, in weighted average and in macro average. The results are the average over the best 5 runs in validation over 10. We report the average and the standard deviation for each metric.
}\label{tab:results_rum_chall}

\vspace{1CM}

\begin{tabular}{| l | l | l | l | l | 
l | l | l | c | c |}
\hline
\textbf{Category} & \textbf{Model} & \textbf{S-F1} & \textbf{Q-F1} & \textbf{D-F1} & \textbf{C-F1} & \textbf{W-F1} & \textbf{M-F1} & \textbf{LR} & \textbf{DO}\\
\hline
\multirow{2}{*}{\textsc{Dummy}}
& \textsc{Maj.} & $0.0\;(\pm0.0)$ & $0.0\;(\pm0.0)$ & $0.0\;(\pm0.0)$ & $82.9\;(\pm0.0)$ & $58.6\;(\pm0.0)$ & \cellcolor{col_table!23} $20.7\;(\pm0.0)$ & $/$& $/$\\
\cline{2-10}
& \textsc{Rand.} & $15.3\;(\pm2.0)$ & $14.4\;(\pm3.4)$ & $11.7\;(\pm2.0)$ & $39.1\;(\pm1.5)$ & $31.8\;(\pm0.7)$ & \cellcolor{col_table!20}$20.1\;(\pm0.8)$ & $/$& $/$\\
\hline
\multirow{2}{*}{\textsc{Basel.}}
& \textsc{Single} & $31.3\;(\pm3.7)$ & $52.5\;(\pm2.6)$ & $27.7\;(\pm5.7)$ & $56.2\;(\pm6.6)$& $51.0\;(\pm5.3)$ & \cellcolor{col_table!35}$42.0\;(\pm3.4)$ & $5.0\cdot 10^{-5}$ & $0.25$ \\
\cline{2-10}
& \textsc{Pair} & $30.2\;(\pm1.5)$ & $54.3\;(\pm1.6)$ & $33.7\;(\pm1.4)$ & $67.2\;(\pm3.7)$ & $59.3\;(\pm2.9)$ & \cellcolor{col_table!60}$46.4\;(\pm1.8)$ &  $2.5\cdot 10^{-5}$ & $0.25$ \\
\hline
\multirow{4}{*}{\textsc{Cont.}} & \textsc{TC} & $28.3\;(\pm3.0)$ & $53.1\;(\pm4.7)$ & $31.1\;(\pm4.2)$ & $68.4\;(\pm5.3)$ & $59.6\;(\pm3.9)$ & \cellcolor{col_table!55}$45.3\;(\pm2.3)$ & $2.5\cdot 10^{-5}$ & $0.5$\\
\cline{2-10}
& \textsc{TC + T} & $27.9\;(\pm1.8)$ & $49.8\;(\pm1.9)$ & $33.6\;(\pm2.9)$ & $63.3\;(\pm4.5)$ & $55.8\;(\pm3.3)$ & \cellcolor{col_table!45}$43.6\;(\pm2.0)$ & $7.5\cdot 10^{-6}$ & $0.25$ \\
\cline{2-10}
& \textsc{TC + U} & $27.9\;(\pm1.1)$ & $52.7\;(\pm2.2)$ & $32.2\;(\pm3.0)$ & $64.8\;(\pm4.0)$ & $57.0\;(\pm3.0)$ & \cellcolor{col_table!50}$44.4\;(\pm1.5)$ & $1.0\cdot 10^{-5}$ & $0.25$ \\
\cline{2-10}
& \textsc{TC + U + T} & $27.2\;(\pm2.1)$ & $51.4\;(\pm3.0)$ & $32.8\;(\pm1.4)$ & $62.2\;(\pm3.2)$ & $55.0\;(\pm2.8)$ & \cellcolor{col_table!44}$43.4\;(\pm2.0)$ & $1.0\cdot 10^{-5}$ & $0.5$ \\
\hline

\end{tabular}
\caption{\textbf{SQDC -  New split.}  F1 scores obtained on the test set of SQDC dataset, with our new split to obtain complex structures even in training. See caption in Table \ref{tab:results_rum_chall} for further details.
}\textbf{\label{tab:results_rum_new}}

\vspace{1CM}

\begin{tabular}{| l | l | l | 
l | l | l | c | c |}
\hline
\textbf{Category} & \textbf{Model} & \textbf{NS-F1} & \textbf{S-F1} & \textbf{W-F1} & \textbf{M-F1} & \textbf{LR} & \textbf{DO}\\
\hline
\multirow{2}{*}{\textsc{Dummy}}
& \textsc{Maj.} & $82.9\;(\pm0.0)$ & $0.0\;(\pm0.0)$ & $58.6\;(\pm0.0)$ & \cellcolor{col_table!10}$41.4\;(\pm0.0)$ & $/$& $/$\\
\cline{2-8}
& \textsc{Rand.} &  $59.6\;(\pm1.0)$ & $38.4\;(\pm1.5)$ & $53.4\;(\pm0.8)$ & \cellcolor{col_table!20}$49.0\;(\pm0.9)$ & $/$& $/$\\
\hline
\multirow{2}{*}{\textsc{Basel.}}
& \textsc{Single} &  $74.4\;(\pm2.9)$ & $52.9\;(\pm0.8)$& $68.1\;(\pm2.3)$ & \cellcolor{col_table!60}$63.6\;(\pm1.8)$ & $1.0\cdot 10^{-5}$ & $0.5$ \\
\cline{2-8}
& \textsc{Pair} & $73.4\;(\pm3.4)$ & $53.8\;(\pm1.5)$ & $67.7\;(\pm2.6)$ & \cellcolor{col_table!60}$63.6\;(\pm2.0)$ &  $7.5\cdot 10^{-6}$ & $0.5$ \\
\hline
\multirow{4}{*}{\textsc{Cont.}} & \textsc{TC} & $73.3\;(\pm3.2)$ & $49.3\;(\pm1.3)$ & $66.3\;(\pm2.5)$ & \cellcolor{col_table!50}$61.3\;(\pm2.1)$ & $7.5\cdot 10^{-6}$ & $0.25$\\
\cline{2-8}
& \textsc{TC + T} & $75.3\;(\pm3.0)$ & $51.1\;(\pm1.4)$ & $68.3\;(\pm2.4)$ & \cellcolor{col_table!57}$63.2\;(\pm2.0)$ & $1.0\cdot 10^{-5}$ & $0.5$ \\
\cline{2-8}
& \textsc{TC + U} & $74.7\;(\pm3.0)$ & $49.9\;(\pm1.0)$ & $67.5\;(\pm2.1)$ & \cellcolor{col_table!54}$62.3\;(\pm1.6)$ & $1.0\cdot 10^{-5}$ & $0.5$ \\
\cline{2-8}
& \textsc{TC + U + T} & $74.7\;(\pm1.5)$ & $48.4\;(\pm1.9)$ & $67.0\;(\pm1.3)$ & \cellcolor{col_table!52} $61.6\;(\pm1.3)$ & $2.5\cdot 10^{-5}$ & $0.25$ \\
\hline

\end{tabular}
\caption{\textbf{SQDC -  Binary.}  F1 scores obtained on the test set of SQDC dataset, with our new split to obtain complex structures even in training, for the binary task to detect Stance class vs No Stance Class. See caption in Table \ref{tab:results_rum_chall} for further details.
}\label{tab:results_rum_bin}

\end{table*}

\begin{figure*}
    \includegraphics[width=\textwidth]{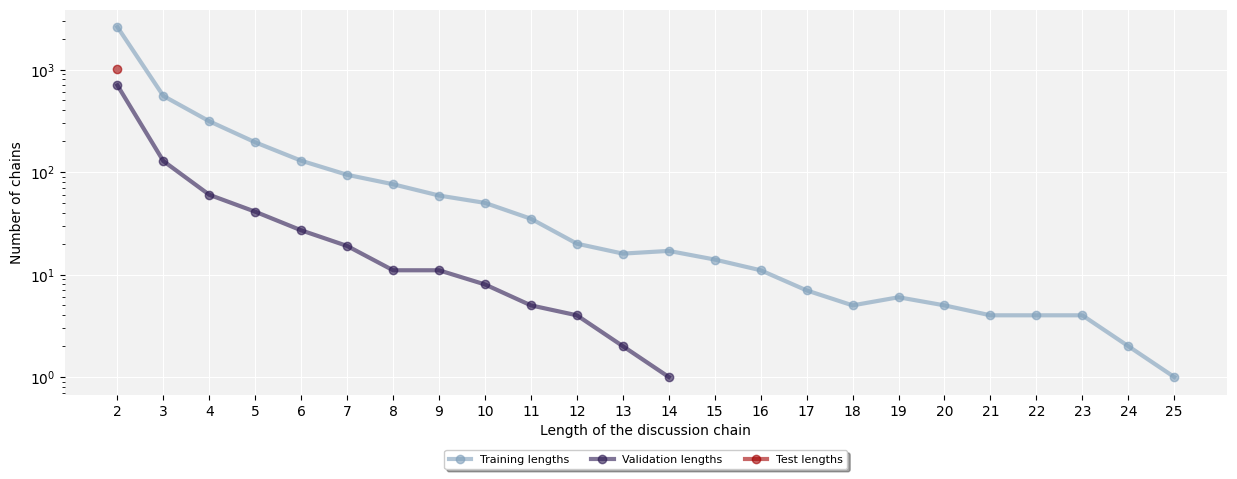}\hfill
    \caption{Length distribution of discussion chains (i.e. number of claims in the discussion chain) in SQDC dataset - challenge version.}\label{fig:length_distribution_chall_rum}

    \vspace{2cm}
    \includegraphics[width=\textwidth]{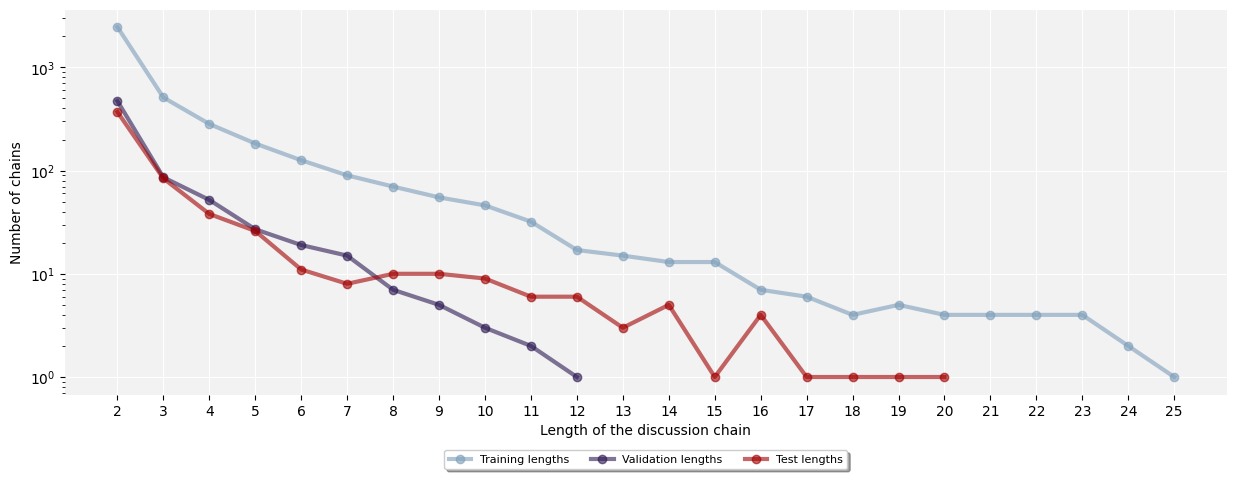}\hfill
    \caption{Length distribution of discussion chains (i.e. number of claims in the discussion chain) in SQDC dataset - new split version.}\label{fig:length_distribution_newsplit_rum}
\end{figure*}

\subsection{Results on SQDC dataset}\label{app:rumoureval}

\textbf{The SQDC dataset.}
We perform the same set of experiments and analysis on a second dataset, which was developed for the task ``SQDC support classification'' at the RumourEval 2019 challenge \citep{gorrell-etal-2019-semeval}. For each item we have the same information as in the SDK dataset, and given a discussion tree, all the discussion chains from the initial claim to any node (even internal) are extracted, and each item labeled according to the last comment. However, the label of each claim does not represent the stance versus the previous claim, but rather the stance with respect to the rumour discussed in the chain. This chain is treated as 
the common ground topic on which the discussion is taking place, even if it is not necessarily stated explicitly in the initial claim. Again, the dataset split is based on the initial claim, avoiding any data contamination.

There are four possible labels: \textsc{i.} \textit{support}, \textsc{ii.} \textit{query}, \textsc{iii.} \textit{deny}, and \textsc{iv.} \textit{comment}. Those labels are respectively shortened as S, Q, D and C, from which the name of the task (SQDC support classification). The original dataset is highly unbalanced among the classes and comprises threads from Reddit\footnote{https://www.reddit.com} and Twitter\footnote{https://twitter.com}. 
We focus this second set of experiments on the Twitter part of the dataset. 

\par \textbf{Experiments.}
 At first, we run our experiments on the original train-validation-test split, reaching different results w.r.t. those obtained on Kialo, since the \textsc{Single} model yields the best performance (see full results in Table \ref{tab:results_rum_chall}). 
\par We further inspect the dataset and we find that the test set was formed only by chains of length $2$, where the usefulness of the context is limited. So, we exclude the original test set and generate a new train-validation-test split, analysing the distribution of labels and chain lengths. 
The results are different w.r.t. the original SQDC dataset: the \textsc{Contextual} model achieves a performance between \textsc{Single} model (lower bound) and \textsc{Pair} model (upper bound). For details, see Table \ref{tab:results_rum_new}. Overall, the  results on the new split of the SQDC dataset confirm the overall findings obtained by analysing the learning curve for different training sizes in Kialo (discussed in Section \ref{sec:second_experiment}): the SQDC dataset is not large enough to allow modelling the context in an effective way. We also try to test our models on a binary task, more similar to stance detection in Kialo, by merging the \textit{query} class, the \textit{deny} class and the \textit{support} class into a unique stance class, and the comment class as a no-stance class. Results are reported in Table \ref{tab:results_rum_bin}. Again, the \textsc{Single} model is the best performing one probably due to the data size and the context does not yield any improvement.

\begin{table}
\small
\centering
\begin{tabular}{| l | c | c | c | c | c | }
\hline
 & \multicolumn{5}{|c|}{\textbf{SQDC Dataset - Challenge}}\\
\hline
\textbf{Set} & \textbf{S} & \textbf{Q} & \textbf{D} & \textbf{C} & \textbf{Total}\\
\hline
Train & $20.2\%$ & $7.9\%$ & $7.6\%$ & $64.3\%$ & $4519$\\
Valid. & $9.0\%$ & $10.1\%$ & $6.8\%$ & $74.1\%$ & $1049$ \\
Test & $13.2\%$ & $5.8\%$ & $8.6\%$ & $72.4\%$ & $1066$\\
\hline
\hline
 & \multicolumn{5}{|c|}{\textbf{SQDC Dataset - New split}}\\
\hline
\textbf{Set} & \textbf{S} & \textbf{Q} & \textbf{D} & \textbf{C} & \textbf{Total}\\
\hline
Train & $13.9\%$ & $8.6\%$ & $7.6\%$ & $69.9\%$ & $3957$\\
Valid. & $12.0\%$ & $8.9\%$ & $8.7\%$ & $70.4\%$ & $689$\\
Test & $11.3\%$ & $10.9\%$ & $7.1\%$ & $70.7\%$ & $595$\\
\hline
\hline
 & \multicolumn{5}{|c|}{\textbf{SQDC Dataset - Binary}}\\
\hline
\textbf{Set} & \multicolumn{2}{|c|}{\textbf{No Stance}} & \multicolumn{2}{|c|}{\textbf{Stance}} & \textbf{Total}\\
\hline
Train & \multicolumn{2}{|c|}{$69.9\%$} & \multicolumn{2}{|c|}{$30.1\%$} & $3957$\\
Valid. & \multicolumn{2}{|c|}{$70.4\%$} & \multicolumn{2}{|c|}{$29.6\%$} & $689$\\
Test & \multicolumn{2}{|c|}{$70.7\%$} & \multicolumn{2}{|c|}{$29.3\%$} & $595$\\
\hline
\end{tabular}
\caption{\label{tab:label-distribution-rumoureval}
Distribution of the labels in SQDC dataset, distinguishing training set, validation set, and test set We report the three versions experiments: chellenge version, new split version and binary version. 
}
\end{table}

For these datasets, we report the descriptive statistics in Table \ref{tab:label-distribution-rumoureval} and plot the length distribution of the discussion chains in Figure \ref{fig:length_distribution_chall_rum} and Figure \ref{fig:length_distribution_newsplit_rum}.

%\vspace{0.5cm}

\par \textbf{Training Details.}
To balance the classes during training, for each epoch we undersample each class in the training set in order to have $s$ samples for each class, where $s$ is the cardinality of the less represented class. We use as loss function the unweighted Cross Entropy. Then, for validation, we use a weighted Cross Entropy Loss according to the cardinality of each class, with weight $w_c = 100/s_c$ for each class, where $s_c$ is the cardinality of the class $c$. We use the same pipeline for hyperparameter optimization and test on fixed hyperparameters  as in SDK dataset (i.e. 5 best runs in validation over 10), performing even the same statistical test. Again, for all the experiments we use a single A40 GPU with 48GB Memory. \newline

\begin{figure*}
    \centering
    \includegraphics[width=\textwidth]{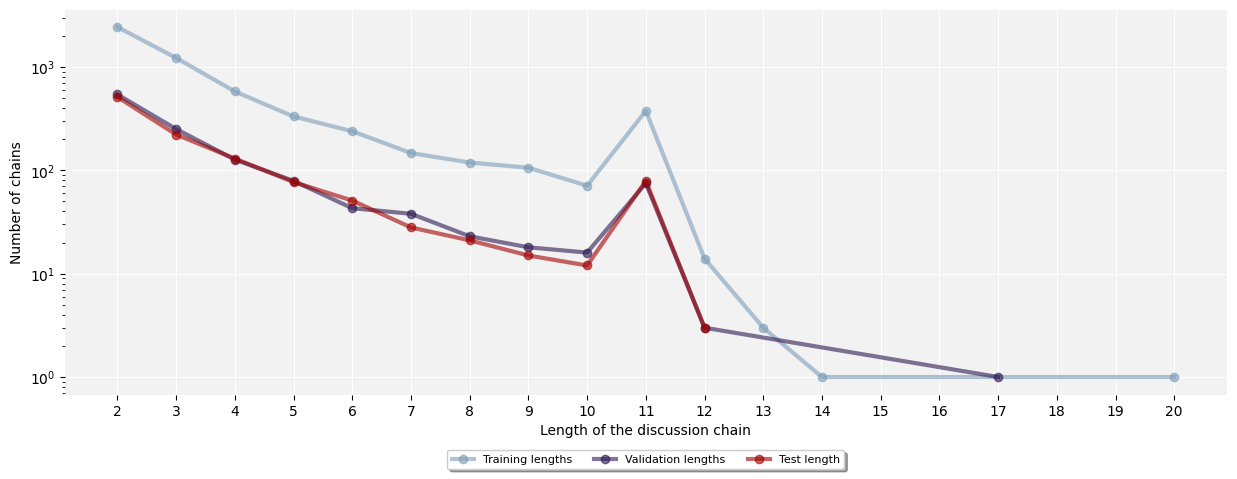}
    \caption{Length distribution of discussion chains (i.e. number of claims in the discussion chain) in ContextAbuse dataset}
    \label{fig:length_distribution_context_abuse}
\end{figure*}

\begin{table*}[!ht]
    \small
    \centering
    \begin{tabular}{|l|l|c|c|c|l|l|l|} \hline 
         \textbf{Category}&  \textbf{Model}&  \textbf{A-F1}&  \textbf{NA-F1}&  \textbf{W-F1}& \textbf{M-F1}& \textbf{LR} & \textbf{DO} \\ \hline 
         \multirow{2}{*}{\textsc{Dummy}} &  \textsc{Majority}&  $89.9 (\pm 0.0)$&  $0.0 (\pm 0.0)$&  $73.4 (\pm 0.0)$& \cellcolor{col_table!20}$45.0 (\pm 0.0)$& /&/ \\ \cline{2-8} 
         &  \textsc{Random}&  $82.2 (\pm 0.4)$&  $21.1 (\pm 2.7)$&  $71.0 (\pm 0.7)$& \cellcolor{col_table!25}$51.7 (\pm 1.4)$& /&/ \\ \hline 
         \textsc{Baselines} &  \textsc{Single}&  $91.0  (\pm 0.4)$&  $70.5 (\pm 0.8)$&  $87.2 (\pm 0.5)$& \cellcolor{col_table!53}$80.7 (\pm 0.6)$& $1.0\cdot 10^{-5}$&$0.5$ \\ \hline 
         \multirow{4}{*}{\textsc{Contextual}} &  \textsc{TC}&  $91.4 (\pm 1.2)$ &  $71.4 (\pm 2.2)$&  $87.7 (\pm 1.3)$& \cellcolor{col_table!55}$81.4 (\pm 1.7)$& $7.5\cdot 10^{-6}$&$0.5$ \\ \cline{2-8} 
 & \textsc{TC + T}& $90.6 (\pm 1.3)$& $69.6 (\pm 2.1)$& $86.7 (\pm 1.5)$& \cellcolor{col_table!52}$80.1 (\pm 1.7)$& $1.0\cdot 10^{-5}$&$0.5$ \\ \cline{2-8} 
 & \textsc{TC + U}& $90.1 (\pm 1.8)$& $68.7 (\pm 2.8)$& $86.2 (\pm 2.0)$& \cellcolor{col_table!50}$79.4 (\pm 2.3)$& $7.5\cdot 10^{-6}$&$0.5$ \\ \cline{2-8} 
         &  \textsc{TC + U + T}&  $91.6 (\pm 0.8)$&  $70.8 (\pm 1.0)$&  $87.8 (\pm 0.8)$& \cellcolor{col_table!54}$81.2 (\pm 0.9)$& $7.5\cdot 10^{-6}$&$0.25$ \\ \hline
    \end{tabular}
    \caption{\textbf{ContextAbuse.} F1 scores obtained on the test set of ContextAbuse dataset. The F1 score is reported for each class, in weighted average and in macro average. The results are the average over the best 5 runs in validation over 10. We report the average and the standard deviation for each metric.}
    \label{tab:contextabuseresult}
\end{table*}

\begin{table}[!ht]
    \small
    \centering
    \begin{tabular}{|l|c|c|c|} \hline  
         &  \multicolumn{3}{|c|}{\textbf{ContextAbuse Dataset}}\\ \hline  
         \textbf{Set}&  \textbf{No Abuse}& \textbf{Abuse} &\textbf{Total}\\ \hline  
         Training&  $82.6\%$&  $17.4\%$&$5651$\\   
         Validation&  $82.4\%$&  $17.6\%$&$1216$\\   
         Test&  $81.7\%$&  $18.3\%$&$1151$\\ \hline 
    \end{tabular}
    \caption{Label distribution in the ContextAbuse dataset}
    \label{tab:contextabusedescr}
\end{table}

\subsection{Results on ContextAbuse dataset}\label{app:contextabuse}

\textbf{The ContextAbuse dataset.}

ContextAbuse \cite{menini2021abuse} is a subset of the well-known hate speech dataset by \citet{Founta_Djouvas_Chatzakou_Leontiadis_Blackburn_Stringhini_Vakali_Sirivianos_Kourtellis_2018}, where the items have been relabeled as "Abusive" or "Not Abusive" taking into account not only the tweet to classify, but also the previous tweets (textual context). This re-annotation led to a remarkable reduction of items annotated as "Abusive", suggesting that context is vital to disambiguate real abusive tweets from other cases (e.g. irony, satire, etc.).
Given the set of tweets from \citet{Founta_Djouvas_Chatzakou_Leontiadis_Blackburn_Stringhini_Vakali_Sirivianos_Kourtellis_2018}, the authors did not retrieve the full discussion tree, but just the discussion chain from the initial claim to the target comment. In this way, there is no overlap among different items, but each tweet in each sequence is seen only once. This could result in major difficulties for contextual models to extract useful information to perform the classification.

\textbf{Experiments}
The dataset is provided on Github\footnote{\url{https://github.com/dhfbk/twitter-abusive-context-dataset/tree/main}}  without official splits. So we create a training/validation/test set according to a 70/15/15 strategy. We report the descriptive statistics in Table \ref{tab:contextabusedescr} and the length of the discussion chain in Figure \ref{fig:length_distribution_context_abuse}. In this case we have only the \textsc{Single} model as a baseline because the goal is to classify a single claim.

The results obtained on the ContextAbuse dataset exhibit similarities to the ones obtained from SQDC dataset (new split version). These findings align with the outcomes of the learning curve experiment from the SDK dataset. In this scenario, the contextual models fail to significantly outperform the baseline (which is the \textsc{Single} model in this case). Nevertheless, it is worth noting that the \textsc{TC} model and \textsc{TC+T+U} model exhibit some improvement, albeit not statistically significant, with the latter showing lower variance. However, it remains uncertain whether, in presence of a larger training set, the contextual model would be capable of increasing the performance gap with the baseline. All the results are reported in Table \ref{tab:contextabuseresult}.

\vspace{0.5cm}
\par \textbf{Training Details.} Differently from the SQDC dataset, for each epoch we use the entire training set without undersampling, and make use of weighted cross-entropy loss both for training loss and validation loss, according to the cardinality of each class (as in Appendix \ref{app:rumoureval}). We use the same pipeline for hyperparameter optimization and test on fixed hyperparameters  as in SDK dataset (i.e. 5 best runs in validation over 10), performing the same statistical test. Again, for all the experiments we use a single A40 GPU with 48GB Memory.

\end{document}